\documentclass{article}

\usepackage{times}
\usepackage{subfigure} 
\usepackage{color}

\usepackage{natbib}

\usepackage{amssymb}
\usepackage{amsmath}
\usepackage{mathtools}
\usepackage{bigdelim}
\usepackage{algorithm}
\usepackage{algorithmic}

\usepackage{hyperref}

\newcommand{\qed}{\nobreak \ifvmode \relax \else
\ifdim\lastskip<1.5em \hskip-\lastskip
\hskip1.5em plus0em minus0.5em \fi \nobreak
\vrule height0.75em width0.5em depth0.25em\fi}

\usepackage[accepted]{icml2016}

\icmltitlerunning{Personalized Risk Scoring for Critical Care Patients}

\begin{document} 
\twocolumn[
\icmltitle{Personalized Risk Scoring for Critical Care Patients using Mixtures of Gaussian Process Experts} 
\icmlauthor{Ahmed M. Alaa$^{1}$}{ahmedmalaa@ucla.edu}
\icmlauthor{Jinsung Yoon$^{1}$}{jsyoon0823@ucla.edu}
\icmlauthor{Scott Hu$^{2}$, MD}{ScottHu@mednet.ucla.edu}
\icmlauthor{Mihaela van der Schaar$^{1}$}{mihaela@ee.ucla.edu}
\icmladdress{$^1$Electrical Engineering Department, University of California, Los Angeles, CA, 90095, USA.\\
$^2$David Geffen School of Medicine, University of California, Los Angeles, CA, 90095, USA.}

\vskip 0.3in
]
\begin{abstract} 
We develop a personalized real-time risk scoring algorithm that provides timely and granular assessments for the clinical acuity of ward patients based on their (temporal) lab tests and vital signs. Heterogeneity of the patients' population is captured via a hierarchical latent class model; the proposed algorithm aims to ``discover" the number of latent classes in the patients' population, and train a mixture of Gaussian Process (GP) experts, where each expert models the physiological data streams associated with a specific class. Self-taught transfer learning is used to transfer the knowledge of latent classes learned from the domain of clinically stable patients to the domain of clinically deteriorating patients. For new patients, the posterior beliefs of all GP experts about the patient's clinical status given her physiological data stream are computed, and a personalized risk score is evaluated as a weighted average of those beliefs, where the weights are learned from the patient's hospital admission information. Experiments on a heterogeneous cohort of 6,313 patients admitted to Ronald Regan UCLA medical center show that our risk score outperforms the currently deployed risk scores, such as MEWS and Rothman scores.   
\end{abstract} 
\section{Introduction}
Risk scoring models that assess the acuity of critical care patients in real-time can guide vital and delay-critical clinical decision-making \cite{churpek2014using}. Unanticipated adverse events such as mortality, cardiopulmonary arrest, or intensive care unit (ICU) transfer are often preceded by disorders in a patient's physiological parameters \cite{kause2004comparison}\cite{hogan2012preventable}. Timely prediction of adverse events can be carried out by continuously quantifying the patient's acuity using evidence in her physiological parameters, and hence assessing her risk for a specific event by computing a real-time ``risk score" that can be tracked by clinicians. Recent systematic reviews have shown that currently deployed expert-based risk scores, such as the MEWS score \cite{morgan1997early}, provide modest contributions to clinical outcomes \cite{tsien1997poor}\cite{cvach2012monitor}\cite{bliss2000behavioural}. Alternatives for expert-based risk scores can be constructed by training a risk scoring model using the data available in electronic medical records \cite{kirkland2013clinical}. Recently, a data-driven risk score, named the Rothman index, has been developed using regression analysis \cite{rothman2013development}. However, current data-driven approaches are mainly constructed using ``one-size-fits-all" models. Personalized models that account for individual traits are anticipated to provide significant accuracy and granularity in risk assessments \cite{snyderman2012personalized}.     
\subsection{Summary of Contributions} 
This paper focuses on the usage of data in the electronic medical records for developing a risk scoring model that provides real-time assessments for the acuity of critical care patients in a hospital ward; such a score can be used to help clinicians make timely ICU admission decisions as an alternative to the currently deployed risk scores in many hospital wards, which are exhibiting high false alarm rates \cite{morgan1997early}\cite{rothman2013development}. 

A major challenge that confronts the development of a reliable risk scoring model is that the patients' population can be highly heterogeneous; patients possess different demographic, lifestyle or clinical features that may affect the interpretability of their physiological stream trajectories, and hence their true risk scores with respect to an adverse event \cite{ng2015personalized}. A risk scoring model trained in a ``one-size-fits-all" fashion may work well on average, but may be consistently underestimating or overestimating the risks for specific patients' sub-populations \cite{snyderman2012personalized}. To that end, {\it personalized medicine} (sometimes referred to as {\it subtyping} \cite{saria2015subtyping}) has emerged as a new approach to medicine that aims at using data to recognize and handle heterogeneity in the patients' populations \cite{chawla2013bringing}.  

The central message of this paper is that ``personalization", in the sense of discovering and accounting for the patients' heterogeneity, can contribute constructively to the clinical utility of the learned risk scoring models. We manifest the clinical significance of personalization by proposing a risk scoring algorithm with the following features:  
\begin{itemize}
\item The algorithm accounts for the patients' heterogeneity through a hierarchical latent class model; it discovers the number of patient classes, and learns a generative Gaussian Process (GP) model for the physiological streams associated with each class using offline training data for clinically stable patients. Discovering the latent classes is carried out using unsupervised learning over the domain of clinically stable patients since they are dominant in the dataset, and are more likely to exhibit stationary physiological trajectories, thus their physiological streams are described with few hyper-parameters.  
\item The knowledge of the latent classes which was extracted from the domain of clinically stable patients is then transferred to the domain of clinically deteriorating patients via {\it self-taught} transfer learning, where the algorithm learns a set of GP models for the different classes of clinically deteriorating patients.  
\item For new patients, the posterior beliefs of all GP experts about the patient's clinical status given her physiological data stream are computed, and the risk score is evaluated as a weighted average of those beliefs. The weights are computed based on the patient's hospital admission information, i.e. the initial admission information for a patient decides the {\it responsibilities} of the different GP experts in assessing her risk. 
\end{itemize}

We report preliminary results for experiments conducted on a cohort of 6,313 patients admitted to Ronald Regan UCLA medical center, and we show that our risk scoring algorithm consistently outperforms baseline algorithms and currently deployed expert-based risk scores. We also highlight the number of GP experts that are learned for the dataset, which reflects the level of heterogeneity of the patient's population, in addition to the admission features that are most relevant to the selection of GP experts.  

\subsection{Related Works}
Various recent works have devoted attention to both the problem of early warning in ICU or hospital wards, and the general problem of personalized risk assessments. In this section, we provide a brief overview of the related previous works, highlighting our contributions and points of departure. 

Modeling physiological streams using GP experts was previously considered in \cite{clifton2012gaussian}, \cite{ghassemi2015multivariate}\cite{durichen2015multitask}, and \cite{pimentel2013modelling}. In all these works, the focus was to predict the values of vital signs via GP regression (e.g. estimating cerebrovascular pressure reactivity in \cite{ghassemi2015multivariate}), and the quality of predictions was assessed using metrics such as the mean-square error. Multi-task GPs were used to model multi-variate physiological streams, and the goal was to construct accurate models for a {\it single} class of patients. That is, inference about a patient's latent class based on the learned GP models for multiple classes of patients was not considered, thus such models can help as an intermediate modeling step for computing risk scores, but cannot be used directly for risk scoring. Moreover, all these models have been limited to the usage of the {\it squared-exponential} covariance kernel (e.g. see eq. (2) in \cite{ghassemi2015multivariate}), which can only captures stationary physiological; for modeling the streams of clinically deteriorating patients (like in the case of patients in hospital ward), we need to consider non-stationary models as well\footnote{Handling non-stationary streams is challenging since in this case we need to learn an unknown number of latent classes, each with an unknown number of states. Moreover, clinically acute patients with non-stationary streams are usually a minority in the training data.}. Finally, all these models are constructed in a ``one-size-fits-all" fashion, i.e. the hyper-parameters are tuned independent of personal and demographic features of the individual patients.   

Apart from GP models, various important advances have been accomplished with respect to personalized and non-personalized risk prognosis problems. In \cite{henry2015targeted} and \cite{dyagilev2015learning}, a Cox regression-based model was used to develop a sepsis shock severity score that can handle data streams that are censored due to interventions. However, this approach does not account for personalization in its severity assessments, and relies heavily on the existence of ordered pairs of comparisons for the extent of disease severity at different times, which may not always be available and cannot be practically obtained from experts. 

Personalized risk models were developed in \cite{ng2015personalized}, \cite{schulam2015framework}, \cite{wang2015towards} and \cite{visweswaran2010learning}. In \cite{ng2015personalized} and \cite{wang2015towards}, personalized risk factors are computed for a new patient by constructing a dataset of $K$ ``similar patients" in the training data, and train a predictive model for that patient. This approach is computationally expensive when applied in real-time for patients in a ward since it requires re-training a model for every new patient, and more importantly, it does not recognize the extent of heterogeneity of the patients, i.e. the constructed dataset has a fixed size of $K$ irrespective to the input distribution. Hence, such methods may incur efficiency loss if $K$ is underestimated, and may perform unnecessary computations if the underlying population is already homogeneous. 

In \cite{schulam2015framework} and \cite{visweswaran2010learning}, Bayesian frameworks to model the patients' clinical status were developed, where a patient's clinical status is assumed to depend on population, sub-population and individual-level parameters. These works share the same Bayesian framework we consider in this paper, and use similar conceptualization for personalization. The main difference between these works and ours is that the disease/disorder is itself a latent variable in our model, and the number of ``subtypes" is unknown. For instance, we do not assume that the whole population in the training set has scleroderma as in \cite{schulam2015framework}, but rather test the infection with scleroderma by learning different models for ``healthy" and ``unhealthy" patients. Since these two populations are usually highly unbalanced, we use transfer learning to transfer the learned heterogeneity (or ``subtyping" as termed in \cite{schulam2015framework}) from one domain to the other.      

\section{Problem Setting}
\subsection{Risk Scoring}
We consider a hospital ward with patients being monitored via temporal physiological streams (e.g. lab tests and vital signs). For every patient $i$, we define $X_{i}(t)$ as a $D$-dimensional stochastic process representing her $D$ physiological streams as a function of time, whereas her $S$ (static) admission features are bundled in a random vector $Y_{i}$. 

Let $v_{i} \in \{0,1\}$ be a binary latent variable that corresponds to the patient's true clinical status; 0 standing for stable patients, and 1 for deteriorating ones. Since physiological streams manifest the patients' clinical statuses, it is natural to assume that the distributional properties of $X_{i}(t)\left|v_{i}=0\right.$ differ from that of $X_{i}(t)\left|v_{i}=1\right.$. A risk scoring model is a function $R: \mathcal{X}_{t}\rightarrow [0,1]$ that maps all possible realizations of the physiological streams up to time $t$, which we denote as $\mathcal{X}_{t}$, to an instantaneous risk score $R(t) \in [0,1]$. The risk score of patient $i$, denoted as $R_{i}(t)$, represents the posterior probability of patient $i$'s clinical status being $v_{i}=1$ having observed samples of $\left(X_{i}(\tau)\right)_{0 \leq \tau \leq t}$. 
\begin{figure}[t!]
    \centering
    \includegraphics[width=3.25 in]{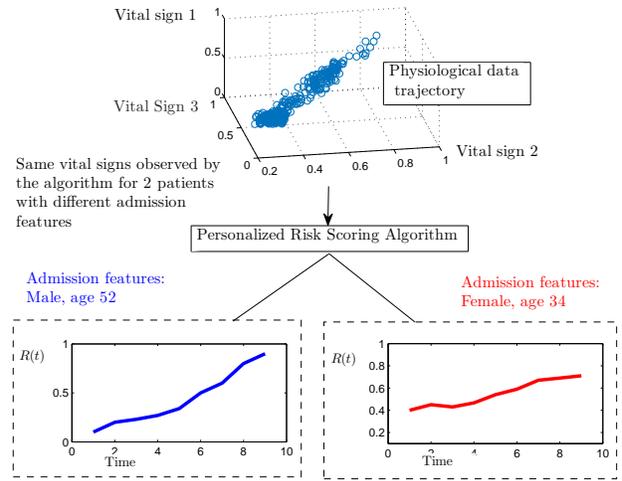}     
    \caption{Depiction of our conceptualization for personalized risk scoring.}
		\label{trj}
\end{figure}
\begin{figure*}[t!]
    \centering
    \includegraphics[width=5.5 in]{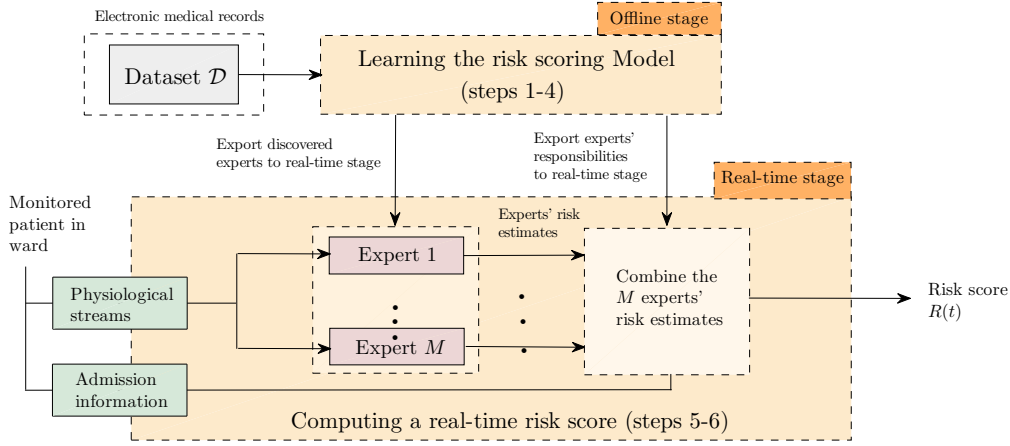}     
    \caption{Block diagram for the proposed risk scoring algorithm.}
		\label{bckdg}
\end{figure*}
\subsection{Objectives}
Given an offline training set $\mathcal{D}$ that comprises $N$ reference patients, we aim to learn a risk scoring model $R: \mathcal{X}_{t}\rightarrow [0,1]$ that best estimates the posterior probability of $v_{i}=1$ for a new patient $i$ after time $t$ (starting from hospital admission) given the observed sequence physiological measurements up to time $t$, and using such an estimate as the risk score $R_{i}(t)$. 

The training dataset $\mathcal{D}$ is represented as a collection of tuples
\[\mathcal{D} = \left\{\left(\left\{X_{i,d}(t_{i,d,n})\right\}_{n=1,d=1}^{N_{i,d},D}, Y_{i}, T_{i},v_{i}\right)\right\}_{i=1}^{N},\] 
where each entry in $\mathcal{D}$ corresponds to a reference patient. The set $\left\{t_{i,d,n}\right\}_{n=1}^{N_{i},D}$ contains the (ordered) time instances at which physiological measurements (e.g. lab tests) for stream $d$ (e.g. blood pressure stream) were gathered for patient $i$, $N_{i,d}$ is the total number of such measurements, $Y_{i}$ is a set of $S$ hospital admission features, $T_{i}$ is the length of patient $i$'s stay in the ward, and $v_{i}$ is her true clinical status: admitted to the ICU or discharged from the ward. The physiological streams are generally irregularly sampled, i.e. $t_{i,d,n}-t_{i,d,n-1}$ is not a constant. We define $\mathcal{D}_{o}$ as the set of all entries in $\mathcal{D}$ for which $v=0$ (i.e. the set of all reference clinically stable patients), and $\mathcal{D}_{1}$ is the set of all entries in $\mathcal{D}$ with $v=1$ (i.e. the set of all reference patients who were admitted to the ICU). The number of patients in dataset $\mathcal{D}_{o}$ is denoted as $N_{o}$. 

We envision a risk scoring model that ensures granularity of its risk assessments by interpreting the physiological streams differently for every individual based on her specific traits. That is, we aim at computing a risk score $R_{i}(t)$ that estimates $\mathbb{P}(v_{i}=1\left|\left(X_{i}(\tau)\right)_{0\leq \tau\leq t}, Y_{i}\right.)$ rather than a one-size-fits-all score that estimates $\mathbb{P}(v_{i}=1\left|\left(X_{i}(\tau)\right)_{0\leq \tau\leq t}\right.)$. In order to learn how to refine the risk assessments, we need to discover the inherent heterogeneity in the data, and learn the relevance of the admission information to this heterogeneity. In the next section, we present an algorithm that accomplishes this task. To illustrate our conceptualization for personalized risk scoring, we plot a physiological stream trajectory in a 3-dimensional space of vital signs in Figure \ref{trj}. This trajectory is simply a trace of normalized vital signs measurements for a certain patient. A one-size-fits-all score would interpret this trajectory as a constant sample path for $R(t)$ for any patient, regardless of her admission information $Y_i$. Contrarily, we target a refined score that interprets such trajectory differently for different patients, i.e. a male of age 52 get different risk assessment than a female of age 34. In order to learn how to refine the risk assessments, we need to discover the inherent heterogeneity in the data, and learn the relevance of the admission information to this heterogeneity. 

\section{The Personalized Risk Scoring Algorithm}
\begin{figure}[h!]
    \centering
    \includegraphics[width=3 in]{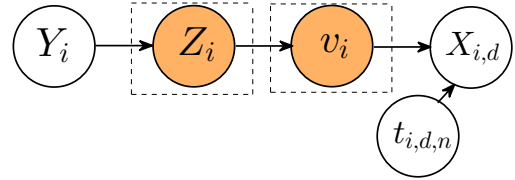}     
    \caption{Graphical model for the generation of vital signs and lab tests (latent variables are shaded).}
		\label{grphm}
\end{figure}

We propose an algorithm for computing a real-time risk score for patients in a hospital ward. As depicted in Figure \ref{bckdg}, the algorithm comprises 6 steps; steps 1-4 are implemented in an offline stage using the data in $\mathcal{D}$, whereas steps 5 and 6 are executed in real-time for new patients. The 6 steps are listed and explained hereunder.\\
\\
{\bf \underline{Offline Stage}:}\\
\\
{\bf \underline{Step 1.} Set up a Hierarchical Latent Class (HLC) model for the physiological streams:} The physiological streams of the patients are modeled as observable variables that are generated via an HLC model \cite{zhang2004hierarchical}. Every patient $i$ possesses a latent variable $Z_{i} \in \{1,2,.\,.\,.,M\}$ that is generated conditional on the admission features $Y_{i}$, and designates her membership in one of $M$ patient classes. The prior on $Z_{i}$ is $\mathbb{P}(Z_{i} = m) = \pi_{m}$. The clinical status $v_{i}$ is generated conditioned on the patient's latent class $Z_{i}$, and based on these latent variables, a physiological stream is generated. A graphical model for the generation of the physiological streams is given in Figure \ref{grphm}. In the remaining steps of the algorithm, our goal is to learn the unknown parameters of this model (including the number of classes $M$), and estimate a personalized risk score
\begin{align}
R_{i}(t) &= \mathbb{P}\left(v_{i}=1\left|\left\{X_{i,d}(t_{i,d,n})\right\}_{n=1,d=1}^{N_{i,d},D}, Y_{i}\right.\right).\nonumber 
\end{align}  

\begin{algorithm}[tb]
   \caption{Personalized Risk Scoring (offline stage)}
   \label{alg:example}
\begin{algorithmic}[1]
   \STATE {\bfseries Input:} Dataset $\mathcal{D}$, precision level $\epsilon$	 
	 \STATE {\bfseries Implement step 2 (Discover the experts):}
	 \STATE Extract dataset $\mathcal{D}_{o}$ of clinically stable patients with label $v=0$
	 \STATE Initialize $M=1$ 
   \REPEAT
	 \STATE $p \leftarrow 1$
   \STATE Initialize $\Theta^{o}_{o,m}$ 
   \REPEAT
   \STATE {\bf E-step:} Compute $Q({\bf \Theta}_{o};{\bf \Theta}^{p-1}_{o}).$
   \STATE {\bf M-step:} ${\bf \Theta}^{p}_{o} = \mbox{arg} \, \mbox{max}_{{\bf \Theta}_{o}} Q({\bf \Theta}_{o};{\bf \Theta}^{p-1}_{o}).$
	 \STATE $Q_{M}^{*} \leftarrow \mbox{max}_{{\bf \Theta}_{o}} Q({\bf \Theta}_{o};{\bf \Theta}^{p-1}_{o}).$
	 \STATE Update responsibilities using Bayes rule $\beta^{p}_{i,m} = \frac{\pi^{p}_{m} \, f_{m}\left(\left.\left\{X_{i,d}(t_{i,d,n})\right\}_{d=1}^{D}\right|\Theta^{p}_{m,o}\right)}{\sum_{m^{'}=1}^{M} \pi^{p}_{m^{'}} \, f_{m^{'}}\left(\left.\left\{X_{i,d}(t_{i,d,n})\right\}_{d=1}^{D}\right|\Theta^{p}_{m^{'},o}\right)}$
   \STATE $p \leftarrow p+1.$
	 \UNTIL{$\frac{1}{N_{o}M}\sum_{i=1}^{N_{o}}\sum_{m=1}^{M}\left|\beta^{p}_{i,m}-\beta^{p-1}_{i,m}\right| < \epsilon$}
	 \STATE $\Psi_{M} = M\left(\frac{D(D+1)}{2}+D+2\right)$
	 \STATE $B_{M,M-1} \approx \frac{\mbox{exp}\left(Q_{M}^{*}-\frac{1}{2}\Psi_{M}\mbox{log}(N_{o})\right)}{\mbox{exp}\left(Q_{M-1}^{*}-\frac{1}{2}\Psi_{M-1}\mbox{log}(N_{o})\right)}$
	 \STATE $M \leftarrow M+1.$
   \UNTIL{$B_{M,M-1} < \bar{B}$}
	\STATE {\bfseries Implement step 3 (Recruit the experts):}
	\STATE Construct the dataset $\left\{Y_{i},{\bf \beta}_{i} = (\beta_{i,o},.\,.\,.,\beta_{i,M})\right\}_{i=1}$.
	\STATE Find linear regression coefficients for $h(Y_{i}) = [w_{1},.\,.\,.,w_{S}]^{T}\,Y_{i}$.
	\STATE {\bfseries Implement step 4 (Self-taught learning):}
	\STATE For every $i \in \mathcal{D}_{1}$ and $m \in \{1,.\,.\,.,M\}$, sample a random variable $c_{i,m} \sim \mbox{Bernoulli}(\beta_{i,m})$.
	\STATE For every expert $m$, construct a dataset $\mathcal{D}_{1,m} = \{i \in \mathcal{D}_{1}: c_{i,m}=1\}$.
	\STATE Find the MLE estimates of $\left\{\Theta_{1,1},.\,.\,.,\Theta_{1,M}\right\}$ using the samples in the corresponding datasets $\left\{\mathcal{D}_{1,1},.\,.\,.,\mathcal{D}_{1,M}\right\}$. 
\end{algorithmic}
\end{algorithm}
{\bf \underline{Step 2.} Discover the Experts through Clinically Stable Patients:} In this step, we aim to recognize the number of patient classes (i.e. the parameter $M$), and estimate a generative model, that we call an ``expert", for the physiological streams of every class, i.e. we target estimating $\mathbb{P}\left(\left.\left\{X_{i,d}(t_{i,d,n})\right\}_{n=1,d=1}^{N_{i,d},D}\right|Z_{i},v_{i}\right)$. We focus on discovering the experts of clinically stable patients since those patients are more likely to exhibit stationary physiological streams (which can be described with fewer hyper-parameters), and are dominant in the dataset $\mathcal{D}$. 

For the entries in $\mathcal{D}_{o}$, we assume that 
\[\left.\left\{X_{i,d}(t_{i,d,n})\right\}_{n=1,d=1}^{N_{i,d},D}\right|Z_{i}=m,v_{i} = 0 \sim \mathcal{GP}(\Theta_{o,m}),\] 
i.e. the physiological streams (as a function of time) are drawn from a GP prior, with a parameter set $\Theta_{o,m}$ for the latent class $Z_{i}=m$, with ${\bf \Theta}_{o} = [\Theta_{o,1},.\,.\,.,\Theta_{0,M}]^{T}$. 

The parameters of the $m^{th}$ expert are given by $\Theta_{o,m} = (\pi_{m}, {\bf M}(t), {\bf K}(t,t^{'})),$ where ${\bf M}(t)$ is a vector-valued mean function ${\bf M}: \mathbb{R}_{+} \rightarrow \mathbb{R}^{D},$ ${\bf K}(t,t^{'})$ is a matrix-valued covariance kernel ${\bf K}: \mathbb{R}_{+}\times\mathbb{R}_{+} \rightarrow \mathbb{R}^{D\times D}$, with ${\bf K} \succeq 0$, and $\pi_{m}$ is the weight of class $m$ in the population. We adopt the stationary multi-task GP model, hence the mean function is a constant vector. The kernel function ${\bf K}(t,t^{'})$ maps the time instances $(t,t^{'}) \in \mathbb{R}^{2}_{+}$ to the matrix $({\bf K}(t,t^{'}))_{d,d^{'}}, \forall d,d^{'} \in \{1,2,.\,.\,.,D\},$ whose entries represent the covariance of the random variables $X_{i,d}(t)$ and $X_{i,d^{'}}(t^{'})$. We assume that the kernel matrix ${\bf K}(t,t^{'})$ has the separable form introduced in \cite{bonilla2007multi} and given by 
\[({\bf K}(t,t^{'}))_{d,d^{'}} = {\bf \Sigma}(d,d^{'})\,k(t,t^{'}),\]
where $k(t,t^{'})$ is a kernel function that quantifies temporal correlations within a stream, and ${\bf \Sigma} \in \mathbb{R}^{D \times D}$ is a {\it cross-stream covariance matrix} that quantifies the correlations between the various physiological streams. In this paper, we will adopt the {\it squared-exponential} covariance kernel \cite{rasmussen2006gaussian}\cite{clifton2013gaussian}\cite{durichen2015multitask}, and we adopt the ``free-form" construction of ${\bf \Sigma}$ via the Cholesky decomposition as follows \cite{bonilla2007multi} 
\[{\bf \Sigma} = {\bf L}\,{\bf L}^{T},\, {\bf L} = \begin{bmatrix}
    \sigma_{1} & 0 & \dots  & 0 \\
		\sigma_{2} & \sigma_{3} & \dots  & 0 \\
    \vdots & \vdots & \ddots & \vdots \\
    \sigma_{\tilde{D}-m+1} & \sigma_{\tilde{D}-m+2} & \dots  & \sigma_{\tilde{D}}
\end{bmatrix},\]
where $\tilde{D} = \frac{D(D+1)}{2}$. Thus, the number of hyper-parameters associated with expert $m$ is $\frac{D(D+1)}{2}+D+2$ \cite{rasmussen2006gaussian}. We denote the multi-variate Gaussian distribution of the random vector $\left.\left\{X_{i,d}(t_{i,d,n})\right\}_{n=1,d=1}^{N_{i,d},D}\right|Z_{i}=m,v_{i}=0$ as $f_{m}\left(\left.\left\{X_{i,d}(t_{i,d,n})\right\}_{n=1,d=1}^{N_{i,d},D}\right|\Theta_{o,m}\right).$

We learn both the number of experts $M$, as well as their parameter sets ${\bf \Theta}_{o}$. This is accomplished through an iterative approach in which we use the expectation-maximization (EM) algorithm for estimating the parameters in ${\bf \Theta}_{o}$ for given values of $M$, and then use the Bayesian information criterion (BIC) to select the number of experts.

The detailed implementation of the EM algorithm is given in lines 4-18 in Algorithm 1. The algorithm is executed by iterating over the values of $M$, with an initial number of experts $M=1$. For every $M$, we implement the usual E-step and M-step of the EM-algorithm: starting from an initial parametrization ${\bf \Theta}_{o}$, in the $p^{th}$ iteration of the EM-algorithm, the auxiliary function $Q({\bf \Theta}_{o};{\bf \Theta}^{p-1}_{o})$ is computed as
\[Q({\bf \Theta}_{o};{\bf \Theta}^{p-1}_{o}) = \mathbb{E}\left[\mbox{log}\left(\mathbb{P}\left(\left. \mathcal{D}_{o},{\bf Z}\right|{\bf \Theta}_{o}\right)\right)\left|\mathcal{D}_{o},{\bf \Theta}^{p-1}_{o}\right.\right],\] 
and the parametrization is updated in the M-step by maximizing $Q({\bf \Theta}_{o};{\bf \Theta}^{p-1}_{o})$ with respect to ${\bf \Theta}_{o}$ (closed-form expressions are available for the jointly Gaussian data in $\mathcal{D}_{o}$ as per the GP model). After that, the algorithm updates {\it expert $m$'s responsibility towards patient $i$} $\beta_{i,m}$ defined as
\[\beta_{i,m} = \mathbb{P}\left(Z_{i}=m\left|\left\{X_{i,d}(t_{i,d,n})\right\}_{n=1,d=1}^{N_{i,d},D}, \Theta_{o,m}\right.\right),\]
i.e. the posterior of patient $i$'s membership in class $m$ given the realization of her physiological data. The iterations of the EM-algorithm stops when the claimed responsibilities of the $M$ experts towards the $N_{o}$ converges to within a precision parameter $\epsilon$.

After each instantiation of the EM-algorithm, we compare the model with $M$ experts to the previous model with $M-1$ experts found in the previous iteration. Comparison is done through the Bayes factor $B_{M,M-1}$ (computed in line 16 via the BIC approximation), which is simply a ratio between Bayesian criteria that trade-off the likelihood of the model being correct with the model complexity (penalty for a model with $M$ experts is given by $\Psi_{M}$ in line 15, such a penalty corresponds to the total number of hyper-parameters in the model with $M$ experts). We stop adding new experts when the Bayes factor $B_{M,M-1}$ drops below a predefined threshold $\bar{B}$.\\ 
\\
{\bf \underline{Step 3.} Recruit the Experts via Transductive Transfer Learning\footnote{Our terminologies with respect to transfer learning paradigms follow those in \cite{pan2010survey}.}:} Having discovered the experts $\left(\mathcal{GP}(\Theta_{o,1}),\mathcal{GP}(\Theta_{o,2}),.\,.\,.,\mathcal{GP}(\Theta_{o,M})\right)$, we need to learn how to associate different experts to the patients based on the initial information we have about them, i.e. the admission features (e.g. transfer status, age, gender, ethnicity, etc). In other words, we aim to learn a mapping rule ${\bf \beta}_{i} = h(Y_{i})$, where ${\bf \beta}_{i} = [\beta_{i,1},.\,.\,.,\beta_{i,M}]^{T}$ is the vector of responsibilities of experts $1$ to $M$ with respect to patient $i$. The values in ${\bf \beta}_{i}$ reflect the extent to which we rely on the different experts when scoring the risk of patient $i$. 

A transductive transfer learning approach is used to learn the function $h(.)$. That is, we use the estimates for the posterior $\beta_{i,m}$ obtained from step 2 (see line 12 in algorithm 1) for every patient $i$ in $\mathcal{D}_{o}$, and then we label the dataset $\mathcal{D}_{o}$ with these posteriors, and transfer these labels to the domain of admission features, thereby constructing a dataset of the form $\left\{Y_{i},{\bf \beta}_{i} = (\beta_{i,o},.\,.\,.,\beta_{i,M})\right\}_{i=1}$. The function $h(.)$ can then be learned via linear regression analysis (see lines 20-21 in Algorithm 1), and the responsibilities of the different experts with respect to any new patient can be assigned by plugging in her admission features in $h(.)$.\\
\\
{\bf \underline{Step 4.} Perform a Self-taught Discovery for the Experts of Clinically Deteriorating Patients:} The knowledge of the $M$ experts about the physiological streams of deteriorating patients is modeled by $\left(\mathcal{GP}(\Theta_{1,1}),\mathcal{GP}(\Theta_{1,2}),.\,.\,.,\mathcal{GP}(\Theta_{1,M})\right)$, and such knowledge needs to be gained from the dataset $\mathcal{D}_{1}$. We use a self-taught transfer learning approach to transfer the knowledge obtained using unsupervised learning from the dataset $\mathcal{D}_{o}$, i.e. the domain of stable patients, to ``label" the dataset $\mathcal{D}_{1}$ and learn the set of experts associated with the clinically acute patients \cite{pan2010survey}\cite{raina2007self}. 

Self-taught learning is implemented by exporting the number of experts $M$ that we estimated from $\mathcal{D}_{o}$ directly to the population of patients in $\mathcal{D}_{1}$, picking a subset of patients in $\mathcal{D}_{1}$ to estimate the parameter set $\Theta_{1,m}$ of expert $m$ by sampling patients from $\mathcal{D}_{1}$ using their responsibility vectors ${\bf \beta}_{i}$ (line 23 in Algorithm 1). The GP model used for the clinically deteriorating patients is a non-stationary model that assigns different hyper-parameters (mean vector and covariance kernel) to different windows of the time domain, and uses the patient ICU admission time $T_{i}$ for every patient $i$ as a surrogate time reference for the physiological stream.\\
\\
\\
{\bf \underline{Real-time Stage}:}\\
\\
{\bf \underline{Step 5.} Consult the Experts:} For a new patient $i$ at time $t$, we consult the $M$ experts about $i$'s clinical status; the knowledge of the experts is represented by the models $\left(\mathcal{GP}(\Theta_{1,1}),.\,.\,.,\mathcal{GP}(\Theta_{1,M})\right)$ and $\left(\mathcal{GP}(\Theta_{o,1}),.\,.\,.,\mathcal{GP}(\Theta_{o,M})\right)$. Consultation is done by computing a set of risk scores $(R_{i,1}(t),R_{i,2}(t),.\,.\,.,R_{i,M}(t))$, where the score $R_{i,m}(t)$ is computed for expert $m$ via the pair of models $(\mathcal{GP}(\Theta_{o,m}),\mathcal{GP}(\Theta_{1,m}))$ by evaluating the posterior probability of the latent variable $v_{i}=1$ given the observed physiological data.\\
\\  
{\bf \underline{Step 6.} Risk Scoring:} An aggregate risk score for patient $i$ is obtained by weighting the opinions of the $M$ experts with their responsibilities ${\bf \beta}_{i}$, which we obtain using the admission features of $i$ using the mapping function $\hat{{\bf \beta}}_{i} = h(Y_{i})$. The risk score of patient $i$ at time $t$ is then given by 
\[R_{i}(t) = \sum_{m=1}^{M}\frac{\hat{\beta}_{i,m}}{\sum_{m^{'}=1}^{M}\hat{\beta}_{i,m^{'}}}\,R_{i,m}(t).\]
Algorithm 2 shows the a pseudo-code for the operations implemented in the real-time stage. 
\begin{algorithm}[tb]
   \caption{Personalized Risk Scoring (real-time stage)}
   \label{alg:example}
\begin{algorithmic}[1]
   \STATE {\bfseries Input:} Physiological measurements $\left\{X_{i,d}(t_{i,d,n})\right\}_{d=1,n=1}^{D,N_{i,d}}$, admission features $Y_{i}$, a set of experts' parameters $\left\{\Theta_{o,1},.\,.\,.,\Theta_{o,M}\right\}$ and $\left\{\Theta_{1,1},.\,.\,.,\Theta_{1,M}\right\}$.
	 \STATE Estimate the experts' responsibilities $\hat{\beta}_{i,m} = w^{T}_{m}Y_{i}$.
	 \STATE For every expert $m$, compute the risk score $R_{i,m}(t) = \mathbb{P}\left(v_{i}=1\left|\left\{X_{i,d}(t_{i,d,n})\right\}_{d=1,n=1}^{D,N_{i,d}},\Theta_{o,m},\Theta_{1,m}\right.\right)$.
   \STATE Aggregate the experts' opinions and compute the final risk score $R_{i}(t) = \sum_{m=1}^{M}\frac{\hat{\beta}_{i,m}}{\sum_{m^{'}=1}^{M}\hat{\beta}_{i,m^{'}}}\,R_{i,m}(t).$ 
\end{algorithmic}
\end{algorithm}

\section{Experiments}
\begin{figure}[t!]
    \centering
    \includegraphics[width=3.25 in]{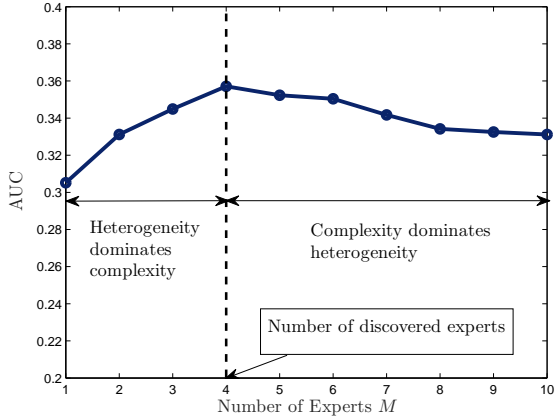}     
    \caption{AUC performance for different number of experts (patient classes) $M$.}
		\label{sm2}
\end{figure}
\begin{figure}[t!]
    \centering
    \includegraphics[width=3.25 in]{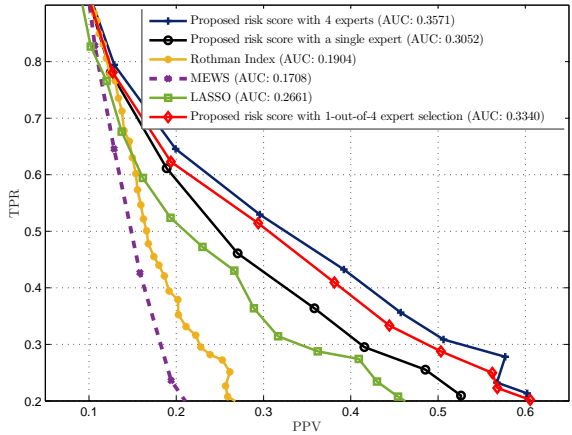}     
    \caption{TPR and PPV performance for the proposed risk scoring algorithm compared to other benchmarks.}
		\label{sm1}
\end{figure}
\subsection{Data description}
Experiments were conducted using a dataset for a cohort of 6,313 patients who were admitted during the years 2013-2016 to a general medicine floor in the Ronald Reagan UCLA medical center, a tertiary medical center. The patient population is heterogeneous: the cohort included patients who were not on immunosuppression and other who were on immunosuppression, including patients that have received solid organ transplantation. In addition, there were some patients that had diagnoses of leukemia and lymphoma. Some of these patients received stem cell transplantation as part of their treatment. Because these patients receive chemotherapy to significantly ablate their immune system prior to stem cell transplantation, they are at an increased risk of clinical deterioration. Of the 6,313 patients, only $8.32\%$ were admitted to the ICU, whereas others were discharged. Thus, $\mathcal{D}$ has 6,313 entries, whereas $\mathcal{D}_{o}$ and $\mathcal{D}_{1}$ have 5,788 and 525 entries respectively. 

Patients in the dataset $\mathcal{D}$ were monitored for 5 vital signs (physiological streams with $D=5$): $O_{2}$ saturation, heart rate, respiratory rate, temperature and systolic blood pressure. The sampling rate for the physiological streams is around 4 hours, and the length of hospital stay for the patients ranged from 2 to 2,762 hours. Each patient is associated with 7 admission features: transfer status, gender, age, race, ethnicity, stem cell transplantation, and admission unit. 

\subsection{Expert Discovery}

For the patient cohort under consideration, the risk scoring algorithm was able to identify 4 patient classes and train the corresponding experts. Figure \ref{sm2} shows the area under curve (AUC) performance of the proposed algorithm versus the number of experts $M$. For $M<4$, the heterogeneity of the patient population dominates the complexity of having many experts over which the training data is split. For $M>4$, adding more experts increases the complexity of the risk model without capturing further heterogeneity, and hence the performance degrades. Picking $M=4$ experts is optimal given the size of $\mathcal{D}$; the algorithm stops after computing the Bayes factor $B_{4,3}$. If the algorithm is to be applied to a larger dataset drawn from the same population, the peak in Figure \ref{sm2} would shift to the right, i.e. more patient classes would be discovered leading to a more granular risk model.  

Having ``discovered the experts", we investigate how the hospital admission features $Y_{i}$ are associated to the responsibility $\beta_{i,m}$ of expert $m$, i.e. we are interested in understanding {\it which} of the admission features are most representative of the latent patient class. Table 1 lists the admission features ranked by their ``importance" in deciding the responsibilities of the 4 experts. The importance, or relevance, of an admission feature is quantified by the weight of that feature $(w_{1},.\,.\,.,w_{S})$ in the learned linear regression function $h(.)$. 

As shown in Table 1, stem cell transplant turned out to be the feature that is most relevant to the assignment of responsibilities among experts. This is consistent with domain knowledge: patients receiving stem cell transplantation are at a higher risk of clinical deterioration due to their severely compromised immune systems, thus it is extremely important to understand their physiological state \cite{hayani2011impact}. This is borne out in table 1 as stem cell transplantation status has the largest contribution in selecting the suitable experts. We note that, in Ronald Regan medical center, patients with leukemia and lymphoma are often taken care of on the same floor as the general medicine population. This then demonstrates the point that it is crucial to utilize information about the heterogeneity of patients to improve their personalized medical care.

Surprisingly, gender turned out to be the second most relevant feature for expert assignments. This means that vital signs and lab tests for males and females should not be interpreted in the same way when scoring the risk of clinical deterioration, i.e. different GP experts needs to handle different genders. The fact that the transfer status of a patient is an important admission factor (ranked third in the list) is consistent with prior studies that demonstrate that patients transferred from outside facilities have a higher acuity with increased mortality \cite{rincon2011association}.  
\begin{table}[h!]
\centering
\caption{\small Relevance of admission features to expert responsibilities.}
\begin{tabular}{l*{2}{c}r}
\hline
{\bf Rank} & {\bf Admission feature} & {\bf Regression coefficient}   \\
\hline
{\bf 1}  & Stem cell transplant & 0.179  \\
\hline
{\bf 2}  & Gender & 0.134 \\
\hline
{\bf 3}  & Transfer status & 0.111 \\
\hline
{\bf 4}  & Race & 0.031 \\
\hline
{\bf 5}  & Ethnicity & 0.021 \\
\hline
\end{tabular}
\end{table}

\begin{figure*}[t!]
    \centering
    \includegraphics[width=6.5 in]{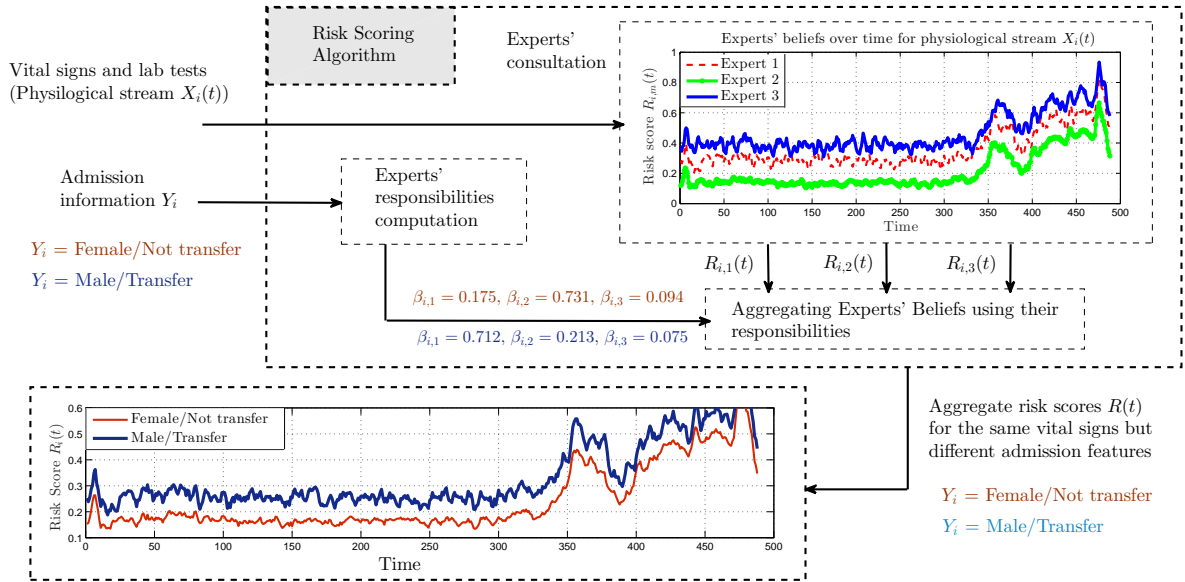}     
    \caption{Depiction for the impact of admission features on expert responsibility assignment.}
		\label{sig2}
\end{figure*}
Figure \ref{sig2} shows traces of the risk signals per expert, and the aggregate risk score over time when the physiological stream applied to the system is the same, but the admission features change. That is, we pick a patient from the dataset $\mathcal{D}$, compute her risk score signal over time, and then do a counter-factual analysis by tracking the risk signal that could have been displayed if instead of the patient being a female who was not transferred from an external clinic was a male who was transferred from another hospital, we see that even if the vital signs are the same, the risk score for the male patient is greater. This is because for the male patient, a larger responsibility score is assigned to expert 1, whose risk signal is higher on average, whereas a female non-transfer patient assigns larger weight to expert 2, who has a less aggressive risk scoring strategy. 

\subsection{Early Warning Performance}
We validated the utility of the proposed risk scoring model by constructing an EWS that issues alarms for ICU admission based on the risk score, and evaluating the performance of the EWS in terms of the positive predictive value (PPV) and the true positive rate (TPR). 

Comparisons for the accuracy of the proposed risk model with two notable state-of-the-art risk scored were carried out, namely: the MEWS score and the Rothman index. The implementation of the MEWS and Rothman indexes followed their standard methodologies in \cite{subbe2001validation} and \cite{rothman2013development}. We also compare the performance of the proposed risk score with the LASSO logistic regression model, the hyper-parameters of which are optimized with respect to the TPR and PPV. All performance measures were computed via a 10-fold stratified cross validation: we ran 10 independent cross validations and reported the average performance for each method. 

As shown in Figure \ref{sm1}, the proposed risk model with $M=4$ experts consistently outperforms the MEWS score, Rothman index and LASSO logistic regression for any setting of the TPR and PPV. The proposed score offers gains of $16\%$ and $18\%$ with respect to the MEWS and Rothman scores ($p$-value $<$ 0.01). This promising results show the value of replacing the currently deployed scores in wards with scores that rigorously learn from heterogeneous data. The value of personalization and recognizing heterogeneity is also manifested in the comparison between our proposed risk model with $M=4$ experts and the same risk model with a single expert. A gain of $5\%$ in the AUC is achieved by accounting for heterogeneity, and a corresponding PPV gain of $10\%$ for a TPR of $40\%$ ($p$-value $<$ 0.01). This means that at a fixed TPR, personalization reduces the rate of false alarms, thereby reducing alarm fatigues and boosting the alarms' credibility (and hence the clinicians' trust in the risk assessment \cite{subbe2001validation}). 

In Figure \ref{sm1}, we also mimic a scenario when our algorithm discovers the experts, and then clinicians use their domain knowledge to select the best expert based only on the stem cell transplantation feature. This corresponds to a ``hard clustering" form of personalization in which patients are exclusively allocated to distinct experts. We see that risk scoring by constructing ``soft clusters" that create mixtures of all experts still outperforms the expert selection approach. The gain would intuitively be larger if the patients' admission features contain a richer set of therapies or interventions. This shows the value of considering all the individual admission features and other latent factors in achieving more granular risk assessments as compared to an expert selection approach that coarsely personalize risk computations.  
        
\section{Conclusion}
We have developed a personalized real-time risk scoring algorithm for critical care patients in hospital wards. The algorithm takes a self-taught approach for learning the heterogeneity of the patients' population: it recognizes heterogeneity in the domain of clinically stable patients, and transfers that knowledge to the domain of clinically acute patients. Heterogeneity is captured via a hierarchical latent class model; a patients' physiological stream is modeled by a GP with hyper-parameters that depend on the patient's latent class. Static hospital admission features are used to construct a customized mixture of GP experts for every new patients, and a risk score is computed by aggregating the posterior beliefs of the different experts. Experiments conducted on a cohort of 6,313 patients demonstrated the superiority of the proposed risk scoring algorithm with respect to currently deployed risk scores, the value of recognizing and accounting for the heterogeneity of patients, and new clinical insights on the relevance of the different admission features to the patients' personalized risk assessment.  
%\nocite{*}
\bibliography{ICU_ref}
\bibliographystyle{icml2016}

%\begin{figure*}[t!]
%    \centering
%    \includegraphics[width=6.5 in]{signal2.eps}     
%    \caption{Depiction for the impact of admission features on expert responsibility assignment.}
%		\label{sig2}
%\end{figure*}

\end{document}